\documentclass[runningheads]{llncs}
\usepackage{soul}
\usepackage[dvipsnames]{xcolor}
\usepackage{graphicx}
\usepackage{babel}
\usepackage{times}
\usepackage{amsmath, amsfonts, amssymb}
\usepackage{caption}
\usepackage{subcaption}
\usepackage{booktabs}
\usepackage{multirow}
\usepackage{tabularx}
\newcolumntype{C}{>{\centering\arraybackslash}X}
\newcommand{\graph}[1]{\mathcal{G}^{(#1)}}
\newcommand{\matrice}[2]{\mathbf{#1}^{(#2)}}

\begin{document}
\title{Maximal Independent Sets for Pooling in Graph  Neural Networks}
%
%
\author{Stevan Stanovic\inst{1}\orcidID{0000-0001-9656-2080} \and
Benoit Gaüzère\inst{2}\orcidID{0000-0001-9980-2641} \and
Luc Brun\inst{1}\orcidID{0000-0002-1658-0527}}
\authorrunning{S. Stanovic et al.}
%
\institute{Normandie Univ, ENSICAEN, CNRS, UNICAEN, GREYC UMR 6072, 14000 Caen, France \\ \email{\{stevan.stanovic, luc.brun\}@ensicaen.fr} \and
INSA Rouen Normandie, Univ Rouen Normandie, Université Le Havre Normandie, Normandie Univ, LITIS UR 4108, F-76000 Rouen, France\\ \email{benoit.gauzere@insa-rouen.fr}}
\maketitle              
\begin{abstract}
    Convolutional Neural Networks (CNNs) have enabled major advances in image classification through convolution and pooling. In particular, image pooling transforms a connected discrete lattice into a reduced lattice with the same connectivity and allows reduction functions to consider all pixels in an image. However, there is no pooling that satisfies these properties for graphs. In fact, traditional graph pooling methods suffer from at least one of the following drawbacks: Graph disconnection or overconnection, low decimation ratio, and deletion of large parts of graphs. In this paper, we present three pooling methods based on the notion of maximal independent sets that avoid these pitfalls. Our experimental results confirm the relevance of maximal independent set constraints for graph pooling.

\end{abstract}
\renewcommand{\thefootnote}{}
    \section{Introduction}
    \label{sec:relatedWork}
        Convolutional Neural Networks (CNNs) achieved major advances in computer
    vision by learning abstract representations of images thought convolution and pooling.  A convolution is a linear filter applied to each pixel of an image which combines its value with the one of its surrounding. The resulting value is usually transformed via a non linear function. The pooling step reduces the size of an image by grouping a connected set of pixels, usually a small squared window, in a single pixel whose value is computed from the ones of window's pixel. Graph Neural Networks (GNNs) take their inspiration from CNNs and aim at transferring advances performed on  images to graphs. However, most of CNNs use images with a fixed structure (shape). While using GNN both the structure of a graph and its content varies from one  graph to another.  Convolution and pooling operations must thus be adapted for graphs. 
    
         A GNN may be defined as a sequence of simple graphs $(\mathcal{G}^{(0)},\dots,\mathcal{G}^{(m)})$ where each $\mathcal{G}^{(l)}=(\mathcal{V}^{(l)}, \mathcal{E}^{(l)})$ is produced by layer $l$ from $\mathcal{G}^{(l-1)}$. Sets  $\mathcal{V}^{(l)}$ and $\mathcal{E}^{(l)}$ denote respectively the set of vertices and the set of edges of the graph. Given $n_{l} = |\mathcal{V}^{(l)}|$, the graph $\mathcal{G}^{(l)}$ may be alternatively defined as $\mathcal{G}^{(l)}=(\textbf{A}^{(l)},\textbf{X}^{(l)})$ where
        $\textbf{A}^{(l)} \in \mathbb{R}^{n_{l} \times n_{l}}$ is the weighted adjacency matrix of  $\mathcal{G}^{(l)}$
        while $\matrice{X}{l} \in \mathbb{R}^{n_{l} \times f_{l}}$ encodes the nodes'  attributes  of  $\graph{l}$ whose dimension is denoted by $f_{l}$. Each line $u$ of $\matrice{X}{l}$ encodes the feature of the vertex $u$ and is   denoted by $x^{(l)}_u$.

        The final graph $G^{(m)}$ of a $GNN$ is usually followed by a Multi-Layer Perceptron (MLP) applied on each vertex for a node prediction task or by a global pooling followed by a MLP for a global graph classification task. 

        \paragraph{Graph convolution.} This operation is mainly realized by a message passing mechanism and allows to learn a new representation for each node by combining the information of the mentioned node and its neighborhood. The neighborhood information is obtained by aggregating all the adjacent nodes information. Therefore, the message passing mechanism can be expressed as follows~\cite{hamilton2020graph}:
        \begin{equation}
            \textbf{x}^{(l+1)}_{u} = UPDATE^{(l)}(\textbf{x}^{(l)}_{u}, AGGREGATE^{(l)}(\{\textbf{x}^{(l)}_{v}, \forall v \in \mathcal{N}(u)\}))\label{eq:message_passing}
        \end{equation}
        where $\mathcal{N}(u)$ is the neighborhood of $u$ and $UPDATE$, $AGGREGATE$ correspond to differentiable functions.

        Let us note that convolution operations should be permutation equivariant, i.e. for any permutation matrix $P\in \{0,1\}^{n_{l} \times n_{l}}$ defined at level $l$, if $f$ denotes the convolution defined at this layer we must have: $f(PX^{(l)}) = Pf(X^{(l)})$.  Note that this last equation, together with equation~\ref{eq:message_passing}, hides the matrix $\textbf{A}^{(l)}$ which nevertheless plays a key role in the definition of the $AGGREGATE$ function by defining the neighborhood of each node.

        \paragraph{Global pooling.} For graph level tasks, a fixed size vector needs to be sent to the MLP. However, due to the variable sizes of graphs within a dataset, global pooling must aggregate the whole graph information into a fixed size vector.  This operation  can be performed by basic operators like  sum, mean or maximum. Let note us that  more complex aggregation strategies~\cite{zhang2018end} also exist. To insure that two isomorphic graphs  have the same representation, global pooling must be  invariant to permutations, i.e. for any permutation matrix $P$, defined at  layer $l$ we must have $g(PX^{(l)}) = g(X^{(l)})$ where $g$ denotes the global pooling operation. 

        \paragraph{Hierarchical pooling.} Summing up a complex graph into a fixed size vector leads necessarily to an important loss of information. The basic idea to attenuate this loss consists in gradually decreasing the size of the input graph thanks to pooling steps inserted between convolution layers. 
        The resulting smaller final graph induces a reduced loss of information in the final global pooling step. This type of method  is called a  hierarchical pooling~\cite{lee2019self,ying2018hierarchical}. 
        The hierarchical pooling step, as the convolution operation should be permutation equivariant in order to keep information localised on desired nodes. Conversely,  global pooling  must be permutation invariant since it computes a graph level representation.
        Let note that, similar to CNNs, the reduced graph leads to a reduction of parameters in the next convolution. However, this reduction is mitigated by the learned part of hierarchical pooling. Moreover, let us consider a line graph with a signal optimally sampled on its vertices. As shown by~\cite{balcilar2021analyzing}, most of GNN correspond to a low pass filter. Applying a GNN on this line graph, hence decreases the maximal frequency of our signal on  vertices producing an over sampling according to the  Nyquist theorem. More details on optimal sampling on graphs may be found in~\cite{anis2014towards,tanaka2020sampling}.

Given a graph $\graph{l}=(\matrice{A}{l},\matrice{X}{l})$ defined at layer $l$     and its reduced version $\graph{l+1}=(\matrice{A}{l+1},\matrice{X}{l+1})$ defined at level $l+1$, the connection between $\graph{l}$ and $\graph{l+1}$ is usually insured by the reduction matrix $\matrice{S}{l} \in \mathbb{R}^{n_l \times n_{l+1}}$ where $n_l$ and $n_{l+1}$ denote respectively the sizes of $\graph {l}$ and $\graph{l+1}$. If $\matrice{S}{l}$ is a binary matrix, each column of $\matrice{S}{l}$ encodes the vertices of $\graph{l}$ which are merged into a single vertex at layer $l+1$. If $\matrice{S}{l}$ is real, each line of $\matrice{S}{l}$ encodes the distribution of each vertex of $\graph{l}$ over the vertices of $\graph{l+1}$. In both cases, we require $\matrice{S}{l}$ to be line-stochastic. 

        Given $\graph{l}=(\matrice{A}{l},\matrice{X}{l})$ and $\matrice{S}{l}$, the feature matrix $\matrice{X}{l+1}$ of $\graph{l+1}$ is defined as follows:
        \begin{equation}
            X^{(l+1)} = S^{(l)\top} X^{(l)}\label{eq:attributes_reduction}
        \end{equation}
        This last equation defines the attribute of each surviving vertex $v_i$ as a weighted sum of the attributes of the vertices $v_j$ of $\graph{l}$ such that  $\matrice{S}{l}_{ji}\neq 0$.
        
        The adjacency matrix of $\graph{l+1}$ is defines by:
        \begin{equation}
            A^{(l+1)} = S^{(l)\top} A^{(l)} S^{(l)}\label{eq:adjacence_reduction}
        \end{equation}
        Let us suppose that $\matrice{S}{l}$ is a binary matrix. Each entry $(i,j)$ of $\matrice{A}{l+1}$ defined by equation~\ref{eq:adjacence_reduction} is equal to $\sum_{r,s}^{n_l} \matrice{A}{l}_{r,s}\matrice{S}{l}_{r,i}\matrice{S}{l}_{s,j}$. Hence two surviving vertices $i$ and $j$ are adjacent in $\graph{l+1}$ if it exists at least two adjacent non surviving vertices $r$ and $s$ such that $r$ is merged onto $i$ ($\matrice{S}{l}_{r,i}=1$) and $s$ onto $j$ ($\matrice{S}{l}_{s,j}=1$). 

        \paragraph{Pooling methods }
        There are two main families of pooling methods. The first family, called Top-$k$ methods~\cite{gao2019graph,lee2019self}, is based on a selection of relevant vertices based on a learned criteria. The second family is based on  node's clustering methods as in DiffPool~\cite{ying2018hierarchical}.
        
         Top-k methods such as gPool~\cite{gao2019graph} learn a score  attached to each vertex by computing the scalar product between the vertex's attributes and one or several learned vectors. Alternatively, a GNN can be used to compute a  relevance vector for each vertex as in SagPool~\cite{lee2019self}. Next, a fixed ratio pooling is used to select the $k$ vertices  with a highest score. Unselected vertices are dropped. In this case, two surviving vertices in the reduced graph will be adjacent only if they were adjacent before the reduction. This last point may result in the creation of disconnected reduced graphs. This disconnection may be avoided by increasing the density of the  graph, using power 2 or 3 of its adjacency matrix or by using the  Kron's reduction~\cite{bianchi2020hierarchical} instead of equation~\ref{eq:adjacence_reduction}. Nevertheless, let us note that simply discarding all non surviving vertices leads to an important loss of information.  We proposed in a previous contribution~\cite{mivspool}, a top-k pooling method called MIVSPool which avoids such drawbacks by using a maximal independent vertex set and graph contraction operations. 

        Clustering based methods learn explicitly or implicitly the matrix $\matrice{S}{l}$ which encodes the reduction of a set of vertices at level $l$ into a single vertex at level $l+1$. Methods (eg.~\cite{ying2018hierarchical}) learning $\matrice{S}{l}$ explicitly have to use a predetermined number of clusters. This last point  forbids the use of graphs of different sizes. Additionally, these methods generally result in dense matrices $\matrice{S}{l}$ which then induce dense adjacency matrices at level $l+1$ (equation~\ref{eq:adjacence_reduction}). As a consequence, graphs produced by these pooling methods have  a density close to 1 (i.e. a  complete graph or an almost complete graph).
        
        An alternative strategy consists in learning $\matrice{S}{l}$ only implicitly. Graph pooling such as the maximal matching method used in EdgePool~\cite{diehl2019towards} may be associated to this strategy. A maximal matching of a graph $\graph{l}=(\mathcal{V}^{(l)},\mathcal{E}^{(l)})$ is a subset $M$ of $\mathcal{E}^{(l)}$, where no two edges are incident to a same vertex, and every edge in $\mathcal{E}^{(l)} \setminus M$ is incident to one of the two endpoints of an edge in $M$.  EdgePool  is based on  a maximal weighted matching technique, i.e. a maximal matching of maximal weight. The weight of each edge, called its score, is learned using the attributes of its two end points. The selected edges are then contracted to form a specific cluster. Note that the use of a maximal weighted matching may result in some vertices not incident to  any selected edges. These vertices are left unchanged. The sequential algorithm~\cite{diehl2019towards} has been parallelized by Landolfi~\cite{landolfi2022revisiting}. Unlike EdgePool, Landolfi~\cite{landolfi2022revisiting} learns a score attached to each vertex and sort all the vertices of the graph according to their score. The weight of each edge is then defined from a combination of the rank of its incident nodes. The similarity between two adjacent vertices is in this case not taken into account.  Moreover, both EdgePool and Landolfi~\cite{landolfi2022revisiting} have a decimation ratio lower than 50\%, which suggests the need for more pooling steps or a poor abstraction in the final graph of the GNN.

        \begin{figure}[t!]
        \centering
        \includegraphics[width=\textwidth]{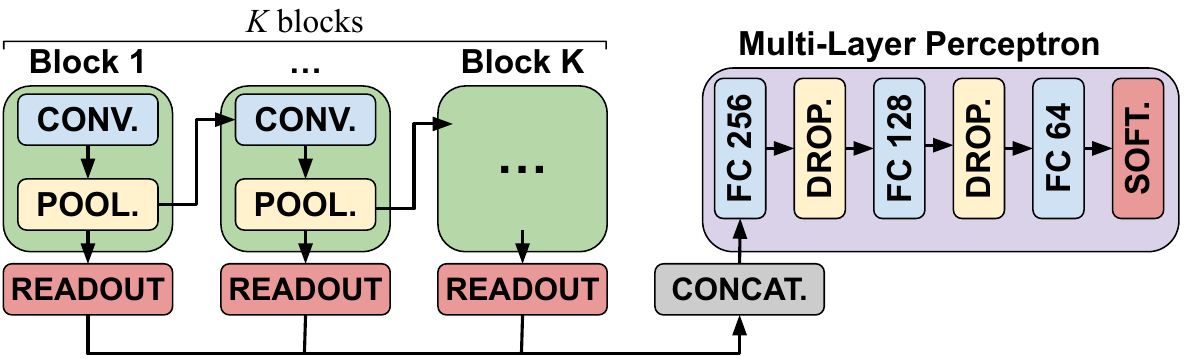}
        \caption{General architecture of our GNN. Each block is composed of a convolution layer followed by a pooling layer. Features learned after each block are sent to the next block and a Readout layer. The \textit{K} vectors resulting from each Readout are concatenated to have several levels of description of the graph and, finally, the concatenation is sent to a Multi-Layer Perceptron.}
        \label{fig:model}
        \end{figure}
        \begin{figure}[t!]
        \centering
        \includegraphics[width=\textwidth]{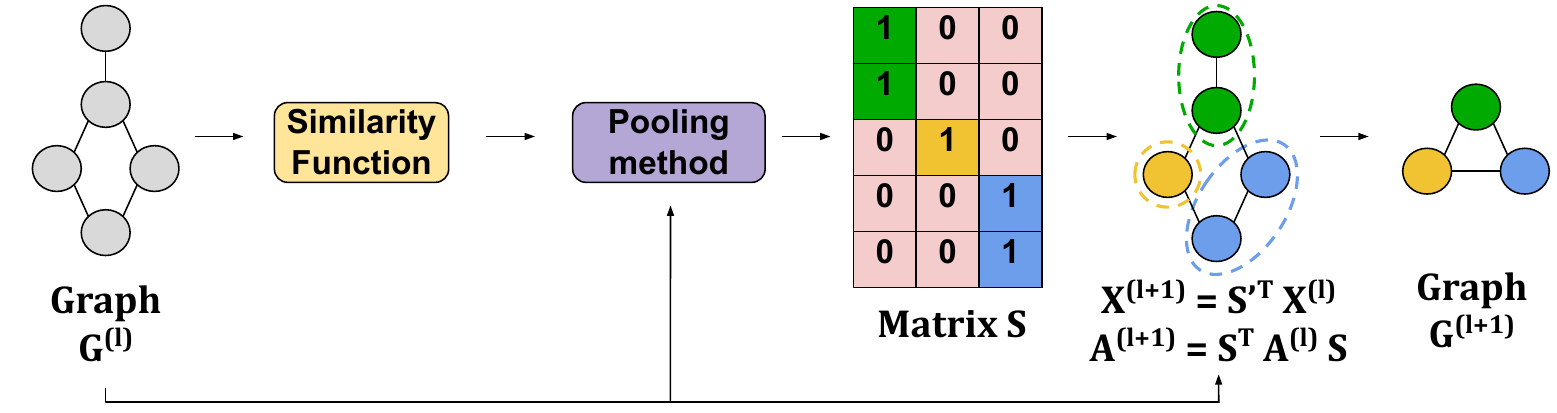}
        \caption{General proposition of our three graph poolings. Each edge is associated to a similarity score (Section~\ref{sec:proposedMethod}). Based on this similarity, a MIS on edge is computed from which a reduction matrix $S$ is derived. Applying $S$ to both feature and structure leads to a reduced graph $G^{(l+1)}$. }
        \label{fig:pooling}
        \end{figure}        

        In this paper, we propose an unified family of graph pooling methods which maintains a  decimation ratio of approximately 50\%, while simultaneously preserving both the structure of the original graph and its attribute information. We achieve this by using a Maximal Independent Set (MIS)~\cite{haxhimusa2007structurally} to select surviving  edges that are evenly distributed throughout the graph, and by assigning non-surviving elements to those that do survive. As a result, we avoid any subsampling or oversampling issues that may arise (see Figure~\ref{fig:pooling}). The source code of the paper is available on the CodeGNN ANR Project Git repository: \url{https://scm.univ-tours.fr/projetspublics/lifat/codegnn}.

    \section{Maximal Independent Sets and Graph Poolings }
        \label{sec:proposedMethod}

        \subsection{Maximal Independent Set (MIS) and Meer's algorithm}
        \label{subsec:mis}
        \textbf{Definition.} Let $\mathcal{X}$ be a finite set and $\mathcal{N}$ a neighborhood function defined on $\mathcal{X}$ such that the neighborhood of each element includes the element itself. A subset $\mathcal{J}$ of $\mathcal{X}$ is a Maximal Independent Set (MIS) if the two following equations are fulfilled:
             \begin{align}
            \forall (x,y) \in \mathcal{J}^{2} &: x \notin \mathcal{N}(y)\label{eq:MIS_first_condition}    \\
            \forall x \in \mathcal{X}-\mathcal{J}, \exists y \in \mathcal{J} &: x \in \mathcal{N}(y)\label{eq:MIS_second_condition}
             \end{align}
        
        The elements of $\mathcal{J}$ are called the surviving elements or survivors. Equations~\eqref{eq:MIS_first_condition} and~\eqref{eq:MIS_second_condition} respectively states that two surviving elements can't be neighbors and each non-surviving element has to be in the neighborhood of at least one element of $\mathcal{J}$. These two equations can be interpreted as a subsampling operation where Equation~\eqref{eq:MIS_first_condition} is a condition preventing the oversampling (two adjacent vertices cannot be selected) while Equation~\eqref{eq:MIS_second_condition} prevents subsampling: Any non-surviving element is at a distance 1 from a surviving one.

       A way to compute a MIS is the Meer's algorithm~\cite{meer1989stochastic} which only involves local computations and is therefore parallelizable. This algorithm attaches a variable to each element. Let us denote by $\mathcal{J}$ the current maximal independent set at an iteration of the algorithm, and let us additionally consider the value $v_x$ attached to an element $x$. Then $x$ is added to $\mathcal{J}$ at current iteration if $v_x$ is maximal among the values of $\mathcal{N}(x)-\mathcal{N(J)}$, where $\mathcal{N(J)}$ denotes $\mathcal{J}$ and its neighbors. Meer's algorithm provides thus a maximal matching such that each of its element is a local maxima at a given step of the algorithm.  We can thus interpret the resulting set as a maximal weight independent set.

        \paragraph{Assignment of non-surviving elements.} After a MIS, $\mathcal{X}$ is split in two subsets: the surviving elements contained in the set $\mathcal{J}$ and the non-surviving elements contained in  $\mathcal{X} - \mathcal{J}$. Simply considering $\mathcal{J}$ as a digest of $X$ may correspond to an important loss of information which simply discards $\mathcal{X}-\mathcal{J}$. In order to avoid such a loss we allow each non surviving element contained in $\mathcal{X}-\mathcal{J}$ to transfer its information to a survivor. The possibility of such a transfer is insured thanks to  equation~\ref{eq:MIS_second_condition} which states that each non surviving element is adjacent to at least one survivor. We can thus associate to any non surviving element $x_j$ a surviving neighbor denoted  by $\sigma(x_j)$.  At layer $l$, the global assignment of non-surviving elements onto surviving ones is encoded by the reduction matrix $\matrice{S}{l} \in \mathbb{R}^{n_l \times n_{l+1}}$ such that :
        
        \begin{equation}
                 \matrice{S}{l}_{ii}=1\quad\forall x_i\in\mathcal{J}\mbox{ and }
                 \matrice{S}{l}_{j\sigma(j)}=1\quad \forall  x_j\in\mathcal{X}-\mathcal{J}
                 \label{eq:construction_S_MIS}
        \end{equation}
        with $\matrice{S}{l}_{ij}=0$ otherwise.
        
        \subsection{Maximal Independent Sets for Graph Pooling}

        \begin{figure}[t]
            \centering
            \mbox{ } \hfill
            \begin{subfigure}{.3\textwidth}
            \centering
            \includegraphics[width=\textwidth]{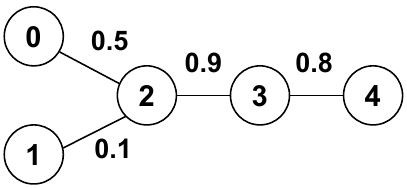}
            \caption{MIES: original graph}
            \label{fig:MIS_a}
            \end{subfigure}
            \hfill
            \begin{subfigure}{.3\textwidth}
            \centering
            \includegraphics[width=\textwidth]{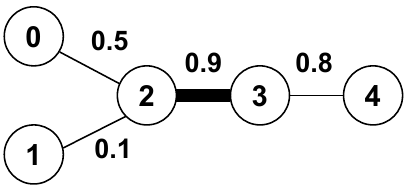}
            \caption{MIES: selection part}
            \label{fig:MIS_b}
            \end{subfigure}
            \hfill
            \begin{subfigure}{.3\textwidth}
            \centering
            \includegraphics[width=\textwidth]{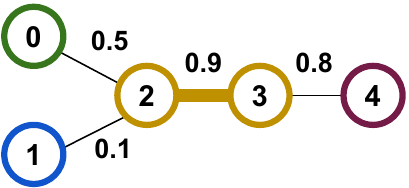}
            \caption{MIES: final clusters}
            \label{fig:MIS_c}
            \end{subfigure}
            
            \hfill\mbox{ }
            \mbox{ } \hfill
            \begin{subfigure}{.3\textwidth}
            \centering
            \includegraphics[width=\textwidth]{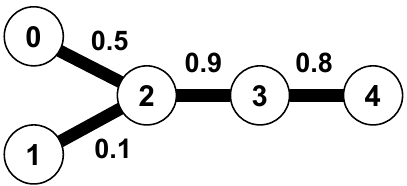}
            \caption{MIESCut: assignment}
            \label{fig:MIS_d}
            \end{subfigure}
            \hfill
            \begin{subfigure}{.3\textwidth}
            \centering
            \includegraphics[width=\textwidth]{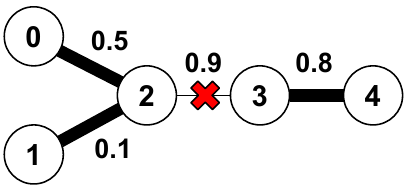}
            \caption{MIESCut: cut}
            \label{fig:MIS_e}
            \end{subfigure}
            \hfill
            \begin{subfigure}{.3\textwidth}
            \centering
            \includegraphics[width=\textwidth]{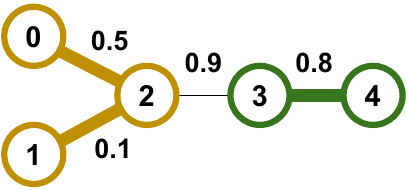}
            \caption{MIESCut: final clusters}
            \label{fig:MIS_f}
            \end{subfigure}
            \hfill\mbox{ }

            \hfill\mbox{ }
            \mbox{ } \hfill
            \begin{subfigure}{.3\textwidth}
            \centering
            \includegraphics[width=\textwidth]{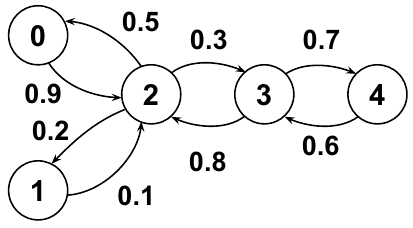}
            \caption{MIDES: original graph}
            \label{fig:MIS_g}
            \end{subfigure}
            \hfill
            \begin{subfigure}{.3\textwidth}
            \centering
            \includegraphics[width=\textwidth]{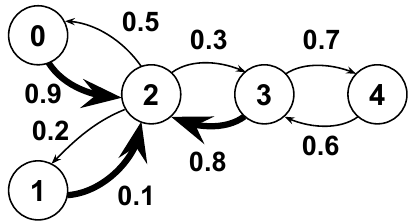}
            \caption{MIDES: selection part}
            \label{fig:MIS_h}
            \end{subfigure}
            \hfill
            \begin{subfigure}{.3\textwidth}
            \centering
            \includegraphics[width=\textwidth]{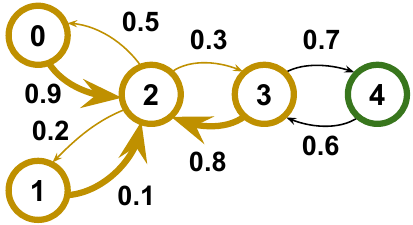}
            \caption{MIDES: final clusters}
            \label{fig:MIS_i}
            \end{subfigure}
            \hfill\mbox{ }
            \caption{Schema of our proposed methods on a toy graph. Number on each edge corresponds to its score $s$ and the bold edges indicates the surviving ones. Each group of vertices with the same color represent a cluster. Figures~\ref{fig:MIS_a} and~\ref{fig:MIS_b} are common steps for MIES and MIESCut.}
            \label{fig:MIS}
        \end{figure} 
        Based on the work~\cite{haxhimusa2007structurally} defined within the image partitioning framework we introduce in the following, three adaptations of these methods in order to define learnable pooling steps. In the following sections,  the adjacency matrix $\matrice{A}{l+1}$ is obtained  from $\matrice{A}{l}$ and  a binary version of $\matrice{S}{l}$ using  equation~\ref{eq:adjacence_reduction}.

        \paragraph{Maximal Independent Edge Set. }
        Most of pooling methods are based on a projection score for each vertex. This strategy is based on the assumption that we can learn  features characterizing relevant vertices for a given classification task. However, even if this hypothesis holds, two adjacent vertices may have  similar scores and the choice of the survivor is in this case arbitrary. An alternative strategy consists in merging similar nodes. Given a GNN with  hierarchical pooling, the graph sequence corresponds to an increasing abstraction from the initial graphs. Consequently,  vertices encoded at each layer of the GNN encode different types of information. Based on this observation, we decided to learn a similarity measure between adjacent vertices at each layer. Inspired by~\cite{verma2018feastnet}, we define the similarity at layer $l$ between two adjacent vertices $u$ and $v$ as $s^{(l)}_{uv} = exp(-\|\matrice{W}{l}.(x_u - x_v)\|)$ where $x_u$ and $x_v$ are the features of vertices $u$ and $v$, $\matrice{W}{l}$ is a learnable matrix and $\|.\|$ is the $L_2$ norm. 

        Given the maximal weighted matching $\mathcal{J}^{(l)}$ defined at level $l$, each vertex of $\graph{l}$ is incident to at most one edge of $\mathcal{J}^{(l)}$. If $u\in \mathcal{V}^{(l)}$ is not incident to $\mathcal{J}^{(l)}$  its features are just duplicated at the next layer. Otherwise, $u$ is incident to one edge $e_{uv}\in \mathcal{J}^{(l)}$ and both $u$ and $v$ are contracted at the next layer. Since $u$ and $v$ are supposed to be similar the attributes of the vertex encoding the contraction of $u$ and $v$ at the next layer must be symmetrical according to $u$ and $v$. To do so, we first define the attribute of $e_{uv}$ as 
        \begin{equation}
            x_{uv}=\frac{1}{2}(x_u^{(l)}+x_v^{(l)})
            \label{eq:edge_attribute}
        \end{equation} 
        where $x_u$ and $x_v$ are the features of vertices $u$ and $v$. The attribute of the merged vertex is then defined as $s_{uv}x_{uv}$.

        An equivalent update of the attributes of the reduced graph may be obtained by computing the matrix $\matrice{S}{l}$ encoding the transformation from $\graph{l}$ to $\graph{l+1}$. This matrix  can  be defined as $\matrice{S}{l}_{ii}=1$ if $i$ is not incident to $\mathcal{J}^{(l)}$, and by selecting arbitrary one survivor among $\{u,v\}$ if $e_{uv}\in \mathcal{J}^{(l)}$. If $u$ is selected we set $\matrice{S}{l}_{uu}=\matrice{S}{l}_{vu}=\frac{1}{2}s_{uv}$. All remaining entries of $\matrice{S}{l}$ are set to $0$.
        Matrix $\matrice{X}{l+1}$ can then be obtained using equation~\ref{eq:attributes_reduction}. We call this method MIESPool and the main steps are presented in Figures~\ref{fig:MIS_a} to~\ref{fig:MIS_c}.

        \paragraph{Maximal Independent Edge Set with Cut (MIESCut). }
        
Graph reduction through maximal weighted matching has two main drawbacks within the GNN framework. First, a maximal matching may produce many vertices not adjacent to the set of selected edges. Such vertices are just copied to the next level which induce a low  decimation ratio (lower than $50\%$). Given that, the number of layers of a GNN is usually fixed, this last drawback may produce a graph with an insufficient level of abstraction at the final layer of the GNN. Secondly, only the score of the selected edges are used to compute the reduced attributes. This last point reduces the number of scores used for the back-propagation  and hence the quality of the learned similarity measures. 

As in the previous section, let us denote by $\mathcal{J}^{(l)}$ the maximal weighted matching defined at layer $l$. By definition of a maximal weighted matching, each vertex not incident to $\mathcal{J}^{(l)}$ is adjacent to at least one vertex which is incident to $\mathcal{J}^{(l)}$. Following~\cite{haxhimusa2007structurally}, we increase the decimation ratio, by attaching isolated vertices to contracted ones. This operation is performed by selecting for each isolated vertex $u$ the edge $e_{uv}$ such that $s_{uv}$ is maximal and $v$ is incident to $\mathcal{J}^{(l)}$. 

This operation provides a spanning forest of $\graph{l}$ composed of isolated edges, trees of depth one (called stars) with one central vertex and paths of length 3. This last type of tree corresponds to a sequence of $4$ vertices with strong similarities between any pair of adjacent vertices along the paths. However, merging all $4$ vertices into a single one, suppose implicitly to apply twice an hypothesis on the transitivity of our similarity measure: more precisely the fact that the two extremities of the paths are similar is not explicitly encoded by our selection of edges. In order to avoid such assumption we remove the central edge of such paths from the selection in order to obtain two isolated edges (see Figures~\ref{fig:MIS_d} to~\ref{fig:MIS_f}).

Let us denote by $\mathcal{J'}^{(l)}$ the resulting set of selected edges which forms a spanning forest of $\graph{l}$ composed of isolated edges and stars.  Concerning the definition of $\matrice{S}{l}$, we apply the same procedure than in the previous section for isolated edges. For stars, we select the central vertex as the surviving vertex. Let us denote by $u$ such a star's center. We then set $\matrice{S}{l}_{uu}=\frac{1}{2}$ and $\matrice{S}{l}_{vu}=\frac{1}{2M}s_{uv}$ for any $v$ such that $e_{uv}\in \mathcal{J'}^{(l)}$ where $M$ is a normalizing factor defined as: $M=\sum_{v|e_{uv}\in \mathcal{J'}^{(l)}} s_{uv}$.  The computation of the attributes of the reduced graph using equation~\ref{eq:attributes_reduction} and matrix $\matrice{S}{l}$ is equivalent to compute for each star's center $u$, the sum, weighted by the score, of the edges' attributes (equation~\ref{eq:edge_attribute}) incident to $u$ and belonging to $\mathcal{J'}^{(l)}$:
\begin{equation}
    x^{(l+1)}_u=\frac{1}{\sum_{v |e_{uv}\in \mathcal{J'}^{l}} s_{uv}}
    \sum_{v |e_{uv}\in \mathcal{J'}^{l}} s_{uv}x_{uv}^{(l)} 
    \label{eq:attributes_MIESCut}
\end{equation}

        \paragraph{Maximal Independent Directed Edge Set. } 
        
        The definition of a spanning forest composed of isolated edges and stars is obtained in three steps by MIESCut: The definition of a maximal weight matching, the attachment of isolated vertices and the cut of all paths of length 3. Following~\cite{haxhimusa2007structurally}, we propose to use the Maximal Independent Directed Edge set (MIDES) reduction scheme which obtains the same type of spanning forest in a single step.  This reduction scheme is based on a decomposition of the edges $e_{uv}$ of the undirected graphs in two oriented edges $e_{u\rightarrow v}$ and $e_{v\rightarrow u}$. The neighborhood of an oriented edge $\mathcal{N}(e_{u\rightarrow v})$ is defined as the union of the sets of edges leaving $u$, arriving on $u$ and leaving $v$. Given  $\graph{l}$ defined at layer $l$ we formally have:
        \begin{equation}
           \mathcal{N}^{(l)}(e_{u\rightarrow v}) =\{e_{u\rightarrow v'}\in\mathcal{E}^{(l)}\}\cup \{e_{v'\rightarrow u}\in\mathcal{E}^{(l)}\}\cup
           \{e_{v\rightarrow v'}\in\mathcal{E}^{(l)}\}
           \label{eq:mides_neighborhood}
        \end{equation}
        The main difference between the neighborhoods defined by equation~\ref{eq:mides_neighborhood} and the one of MIES is that we do not include in the neighborhood edges arriving on $v$. This asymmetry allows the creation of stars centered on $v$. The MIDES algorithm computes a MIS on the edge set using the neighborhood defined by~\eqref{eq:mides_neighborhood} (see Figures~\ref{fig:MIS_g} to~\ref{fig:MIS_i}).

        At layer $l$, applying a MIDES on $\graph{l}$ requires to define a score function on directed edges. We propose to use $s_{uv} = exp(-\|W.(x_u - x_v) + b\|)$ where the bias term $b$ allows to introduce an asymmetry so that $s_{uv}\neq s_{vu}$ if $x_u\neq x_v$. 
        
        Let us denote by $\mathcal{D}^{(l)}$ the set of directed edges produced by a MIDES on $\graph{l}$ using our scoring function. The set $\mathcal{D}^{(l)}$ defines on $\graph{l}$ a spanning forest composed of isolated vertices, isolated edges and stars~\cite{haxhimusa2007structurally}.
        
        For an isolated vertex $u$ we duplicate this vertex at the next layer and copy its attributes. We thus set $\matrice{S}{l}_{uu}=1$. 
        
        For an isolated directed edge $e_{u\rightarrow v}\in \mathcal{D}^{(l)}$ we select $v$ as a surviving vertex and set $\matrice{S}{l}_{vv}=\frac{s_{uv}}{M}$ and  $\matrice{S}{l}_{uv}=\frac{s_{vu}}{M}$ where $M=s_{uv}+s_{vu}$.  This setting corresponds to the following update of the attributes: $x^{(l+1)}_v=(s_{uv}.x^{(l)}_v + s_{vu}.x^{(l)}_u)/(s_{uv}+s_{vu})$. Let us note that since $e_{u\rightarrow v}\in  \mathcal{D}^{(l)}$ we have $s_{uv}>s_{vu}$. The previous formula put thus more weight on the surviving vertex $v$. This update may be considered as a generalization of equation~\ref{eq:edge_attribute} using the asymmetric scores $s_{uv}$ and $s_{vu}$.

        A star within the MIDES framework is defined by a set of edges $e_{w\rightarrow v}$ of $\mathcal{D}^{(l)}$ arriving on the same vertex $v$. We then set $v$ as survivor and generalize the update of the attributes defined for isolated edges by setting $\matrice{S}{l}_{vv}=\frac{1}{N}\sum_{u|e_{u\rightarrow v}\in \mathcal{D}^{(l)}}\frac{s_{uv}}{M_u}$ and $\matrice{S}{l}_{uv}=\frac{1}{N}\frac{s_{vu}}{M_{u}}$ for all $u$ such that $e_{u\rightarrow v} \in \mathcal{D}^{(l)}$ where $M_u=s_{uv}+s_{vu}$ and $N$ is the number of such $u$. Such a definition of $\matrice{S}{l}$ is equivalent to set the updated attribute of $v$ as the mean value of its incident selected edges:
    \[
    x^{(l+1)}_v=\frac{1}{N}\sum_{u|e_{u\rightarrow v} \in \mathcal{D}^{(l)}} \frac{s_{uv}x^{(l)}_v+s_{vu}x^{(l)}_u}{s_{uv}+s_{vu}}\mbox{ with } N=|\{u \in \mathcal{V}^{(l)}| e_{u\rightarrow v}\in \mathcal{D}^{(l)}|.
    \]

    \begin{table}[b!]
            \centering
            \begin{tabularx}{.8\textwidth}{lCCCC}\toprule
                \textbf{Dataset} & \textbf{\#Graphs} & \textbf{\#Classes} & \textbf{Avg $|\mathcal{V}|$} & \textbf{Avg $|\mathcal{E}|$} \\
                \hline
                D\&D~\cite{dobson2003distinguishing} & $1178$ & $2$ & $284 \pm 272$ & $715 \pm 694$ \\
                REDDIT-BINARY~\cite{reddit} & $2000$ & $2$ & $430 \pm 554$ & $498 \pm 623$\\
                \bottomrule
            \end{tabularx}
            \caption{Statistics of datasets}
            \label{tab:dataset}
        \end{table}
        
    \section{Experiments}
         \label{sec:xp}
        
         \paragraph{Datasets. }
         We evaluate our contribution to a bio-informatics  and a social dataset called respectively D\&D~\cite{dobson2003distinguishing} and REDDIT-BINARY~\cite{reddit} whose  statistics are reported on Table~\ref{tab:dataset}. The aim of D\&D is to classify proteins as either enzyme or non-enzyme. Nodes represent the amino acids and two nodes are connected by an edge if they are less than 6 \AA{}ngström apart. REDDIT-BINARY is composed of graphs corresponding to online discussions on Reddit. In each graph, nodes represent users, and there is an edge between them if at least one of them respond to the other’s comment. A graph is labeled according to whether it belongs to a question/answer-based community or a discussion-based community.

        \begin{table}[t!]
                \centering
                \begin{tabularx}{\textwidth}{lCC}\toprule
                    \textbf{Methods} & \textbf{D\&D~\cite{dobson2003distinguishing}} & \textbf{REDDIT-BINARY~\cite{reddit}}\\
                    \hline
                    Baseline &   $76.29 \pm 2.33$ & $87.07 \pm 4.72$\\
                    gPool~\cite{gao2019graph} & $75.61 \pm 2.74$ & $84.37 \pm 7.82$\\
                    SagPool~\cite{lee2019self} & $76.15 \pm 2.88$ & $85.63 \pm 6.26$\\
                    EdgePool~\cite{diehl2019towards} & $72.59 \pm 3.59$ & $87.25 \pm 4.78$\\
                    MIVSPool~\cite{mivspool} & $76.35 \pm 2.09$ & $\mathbf{88.73 \pm 4.43}$\\
                    MIESPool & \textcolor{blue}{$77.17 \pm 2.33$} & $88.08 \pm 4.55$\\
                    MIESCutPool & $\mathbf{77.74 \pm 2.85}$ & $86.47 \pm 4.57$\\
                    MIDESPool & $76.52 \pm 2.21$ & \textcolor{blue}{$88.40 \pm 4.74$} \\
                    \bottomrule
                \end{tabularx}
               \caption{Average classification accuracies obtained by different pooling methods. Highest and second highest accuracies are respectively in \textbf{bold} and \textcolor{blue}{blue}. $\pm$ indicates the $95 \%$ confidence interval of classification accuracy.}
                \label{tab:result}
        \end{table}
        
        \paragraph{Model Architecture and Training Procedure. }
        Our model architecture is composed of $K$ blocks where each block consists of a GCN~\cite{kipf2017semi} convolution layer followed by a pooling layer. The vector resulting of each pooling operation is then sent to the next block (if it exists) and a Readout layer. A Readout layer concatenates the average and the maximum of vertices' features matrix $\matrice{X}{l}$ and these $K$ concatenations are themselves concatenated and sent to a Multi-Layer Perceptron (MLP). The MLP is composed of three fully connected layers and a dropout is applied between each of them. Finally, a Softmax layer is used to determine the binary class of graphs. Note that no batch normalization is applied (Figure~\ref{fig:model}).
        
        To evaluate our model, we use the training procedure proposed by~\cite{fair_comparison}. This procedure performs an outer 10-fold cross-validation (CV) to split the dataset into ten training and test sets. For each outer fold, another 10-fold CV (inner) is applied to the training set to select the best hyperparameter configuration. 
        Concerning hyperparameters, learning rate is set to $10^{-3}$,  weight decay to $10^{-4}$  and batch size to $512$. Other hyperparameters are tuned using a grid search to find the best configuration. Possible values for the  hidden layers sizes are $\left\{64,128\right\}$,  dropout ratio is chosen within $\left\{0.2,0.5\right\}$ and the number of blocks $K$ between 1 and 5. We use the Adam optimizer and maximal number of epochs is set to  $1000$ with an early stopping strategy  if the validation loss has not been improved 100 epochs. For EdgePool, due to time constraints, we fixed the hidden layers sizes at 128 and the dropout ratio at 0.5.
         
        We compare, in Table~\ref{tab:result},  our methods to five state-of-art methods: Baseline ($K$ blocks of GCN~\cite{kipf2017semi}), gPool~\cite{gao2019graph}, SagPool~\cite{lee2019self}, EdgePool~\cite{diehl2019towards} and MIVSPool~\cite{mivspool}, our previous MIS method. 
        First, we note that the baseline obtains quite good results while not implementing any pooling strategy. It highlights the fact that defining a good pooling operation is not trivial. State-of-the-art methods mostly fail at this task,  certainly due to the  significant loss of information resulting from the hard selection of surviving vertices using a top$-k$ strategy. This hypothesis is confirmed by the better results obtained by MIVSPool. Let us note also that for D\& D, based on T-tests with a significance level of 5\%, the average accuracy of EdgePool is statistically lower than the ones of MIS methods.
        Second, we can observe that the strategies combining edge selection methods and MIS (MIESPool, MIESCutPool, MIDESPool) achieve either the highest or the second highest performances. This empirical results tend to demonstrate that the selection on edges may be most relevant, and that a MIS strategy improves the effectiveness of the pooling over EdgePool.
        Finally, best results are obtained by different MIS strategies, hence indicating that the right MIS strategy may be dataset dependant. This hypothesis has to be tested using more extensive hyperparameters selection.

    \section{Conclusion}
    Graph poolings  based on Maximal Independent Sets (MIS) allow, unlike state-of-art methods, to maintain a fixed decimation ratio close to 50\%, to preserve  vertex information and to avoid subsampling and oversampling. Results obtained by our three methods based on MIS confirm the interest of this approach but further investigations on other datasets are needed to conclude on the effectiveness of our methods. The design of alternative similarity scores also corresponds to a promising line of research.

    \paragraph{Acknowledgements:} The work reported in this paper was supported by French ANR grant \#ANR-21-CE23-0025 CoDeGNN and  was performed using HPC resources from GENCI-IDRIS (Grant 2022-AD011013595) and computing resources of CRIANN (Grant 2022001, Normandy, France).

%
%
%
    \bibliographystyle{splncs04}
    \bibliography{bibliography}

\end{document}